\documentclass[11pt,letterpaper]{article}

\usepackage[utf8]{inputenc}
\usepackage[T1]{fontenc}
\usepackage{amsmath,amsfonts,amssymb}
\usepackage{graphicx}
\usepackage{float}
\usepackage{subcaption}
\usepackage{booktabs}
\usepackage{multirow}
\usepackage{url}
\usepackage{hyperref}
\usepackage{xcolor}
\usepackage{listings}
\usepackage[margin=1in]{geometry}
\usepackage{natbib}

\usepackage[T2A]{fontenc}
\usepackage[russian,english]{babel}  

\usepackage{CJKutf8}

\hypersetup{
    colorlinks=true,
    linkcolor=blue,
    citecolor=black,
    urlcolor=blue
}

\newcommand{\textjapanese}[1]{\begin{CJK}{UTF8}{min}#1\end{CJK}}

\title{Anatomy of an Idiom: \\
Tracing Non-Compositionality in Language Models}

\stepcounter{footnote}  

\author{\\ \textbf{Andrew Gomes}\thanks{Correspondence: \texttt{andrew.gomes@epfl.ch}} \\
Theoretical Particle Physics Laboratory, \\
Institute of Physics, EPFL, Lausanne, Switzerland
}

\date{}

\begin{document}

\maketitle

\begin{abstract}
We investigate the processing of idiomatic expressions in transformer-based language models using a novel set of techniques for circuit discovery and analysis. First discovering circuits via a modified path patching algorithm, we find that idiom processing exhibits distinct computational patterns. We identify and investigate ``Idiom Heads,'' attention heads that frequently activate across different idioms, as well as enhanced attention between idiom tokens due to earlier processing, which we term ``augmented reception.'' We analyze these phenomena and the general features of the discovered circuits as mechanisms by which transformers balance computational efficiency and robustness. Finally, these findings provide insights into how transformers handle non-compositional language and suggest pathways for understanding the processing of more complex grammatical constructions.
\end{abstract}

\section{Introduction}

The field of mechanistic interpretability has made significant progress in recent years towards understanding the internal workings neural networks, in particular those of transformers \citep{vaswani2017attention} designed for natural language production. See \citet{rai2024practical} and \citet{sharkey2025open} for recent reviews. Human interpretability of the internal model computations is not typically a requirement of the training process, and is therefore not a manifest feature of trained language models. Nevertheless, a number of interpretability tools have become available, including (but not limited to) autoencoders and transcoders \citep{bricken2023towards, cunningham2023sparse,yun2021transformer,dunefsky2024transcoders, ameisen2025circuit} and circuit identification using activation patching \citep{meng2022locating, zhang2023best, conmy2023acdc}, path patching \citep{wang2022interpretability, goldowsky2023localizing}, and contextual decomposition \citep{hsu2024efficient}.

Of particular interest is the way in which transformers process natural language in order to accomplish whatever task they are assigned (e.g., next word prediction). Take for example the case of \citet{wang2022interpretability}, where the authors discovered a circuit responsible for indirect object identification (IOI). The components of the circuit included the previously identified Induction Heads \citep{elhage2021mathematical, olsson2022context}, used to copy context information, as well as newly discovered Backup Heads, providing robustness to circuit failures. Very roughly, the transformer's task can be split into two parts: understanding the context and problem (early in the network), and producing the correct answer (later in network). In this paper, we focus on the former: what mechanisms do transformers use to process and make sense of natural language?

A particular feature of language is its inherent non-compositionality: constructions where the whole meaning cannot be inferred from its parts. Idiomatic expressions are the prototypical example of such constructions, as they require language models to override a literal meaning in favor of a figurative one. For example, consider the sentence \textit{That was a piece of cake}. An English speaker understands that this expression means \textit{That was easy}, despite the literal meanings of the individual words. Prior work has revealed that language models recognize idioms \citep{tan-jiang-2021-bert, 10.3389/frai.2022.813967}, store idioms in the model weights of early layers \citep{miletic-walde-2024-semantics, haviv2023understanding}, and treat them much like single lexical units \citep{dankers-etal-2022-transformer}. However, the precise mechanisms by which transformers process idioms remain relatively unexplored and their determination represents a crucial first step toward understanding the processing of more complex grammatical and semantic phenomena.

In this work we use a modified path patching algorithm to discover circuits responsible for the processing common English idioms by Gemma 2--2B \citep{gemma2024}, though none of our findings depend on the particular architecture of this model. Through systematic analysis of these circuits, we present three novel findings about idiom processing in Gemma 2--2B:

\begin{enumerate}
\item \textbf{Non-compositional semantic processing}: Idioms take on their non-compositional (figurative) meaning in the early layers of the network, mirroring previous work. Moreover, this processing exhibits a two-phase structure, with cross-token attention in early layers (0--2) followed by semantic integration in later layers (3--5). Most computation, even between non-final idiom elements, occurs on the final token of each expression.

\item \textbf{Specialized Idiom Heads}: We identify specific attention heads that consistently activate across multiple idioms, suggesting functional specialization for non-compositional language processing. Nevertheless, each idiom employs a distinct direction in Query-Key space rather than a universal ``idiom direction.''\footnote{After the completion of this work, the recent preprint \citet{oh2025tug} came to our attention in which they identify ``idiomatic heads'' in Llama3.2--1B via a different causal technique. They focus primarily on how transformers disambiguate between the literal and figurative interpretations of idioms for the purpose of next-token generation, with less focus on the properties of these heads and mechanisms involved in the semantic processing of idioms themselves.}

\item \textbf{Augmented reception}: Early processing of idiom tokens creates enhanced receptivity of these tokens to each other in later attention layers, pointing to a mechanism by which transformers efficiently allocate attention.
\end{enumerate}

We frame these features as mechanisms by which transformers navigate the tension between computational efficiency and robustness in language processing. Our findings provide novel insights into how transformers handle non-compositional semantics and establish a framework for investigating the processing more complex linguistic constructions.

The remainder of this paper is organized as follows: Section~\ref{sec:circuit-creation} describes our circuit discovery methodology, starting with idiom selection and corruption strategies. After reviewing the ACDC (activation patching circuit discovery) framework, we introduce a novel path patching algorithm. Section~\ref{sec:results} presents our results, analyzing circuit properties, Idiom Heads, Query-Key space representations, and augmented reception mechanisms. We conclude in Section~\ref{sec:discussion} with a discussion of the implications for non-compositional language processing and broader transformer mechanisms.

\section{Circuit Discovery}
\label{sec:circuit-creation}

In this section we outline the methodology for constructing idiom processing circuits. We employ a modified ACDC framework \citep{conmy2023acdc} (note that ACDC employs activation patching as distinguished from path patching) to identify minimal computational circuits. One can divide the model operations into two parts: those acting between the tokens of an idiom, and those involved in the processing of information on a single token. The latter is exemplified by the MLP layer, but also includes a token attending to itself or to earlier tokens not directly a part of the idiom (e.g., the \textit{<BOS>} token). These ``single-token'' operations are crucial to idiom processing, for example converting semantically literal representational data into figurative representational data. How this occurs is an important direction of research (e.g., \citet{meng2022locating}), however, in this work we will focus on the sufficiently complex problem of understanding the former ``inter-token'' operations.

\subsection{Step 1: Idiom Selection and Corruption}

We first select a common English idiom and its associated figurative meaning, e.g., \textit{a piece of cake} (\textit{easy}), \textit{kicked the bucket} (\textit{died}), \textit{hit the sack} (\textit{went to sleep}), etc. For each idiom, we generate corrupted versions by replacing key tokens (which we call the ``original tokens,'' e.g., \textit{piece}) with ``replacement tokens'' (e.g., \textit{slice}).\footnote{In general, tokens include a space before the word, a feature of the Gemma 2 tokenizer. However, for ease of reading we will refer to them without the space.} In order for this corruption to tell us something about the idiom's processing, we require that the embedding vector of the replacement token should have high cosine similarity to the original token (see Appendix~\ref{app:corruptions} for an example list of candidates), preserve the syntactic structure of the idiom, and only minimally change the \textbf{literal} semantic meaning. On the other hand, a corruption should not carry a similar \textbf{figurative} semantic meaning. For example, one could replace \textit{kicked} in \textit{kicked the bucket} with \textit{booted}, but not with \textit{kicks} (which would similarly mean \textit{died}). The reason for these requirements will become clear later.

To evaluate the effectiveness of both the figurative meaning and the corruptions in making sense of the idiom, we first convert them into the beginnings of sentences in a uniform way: \textit{That was a piece of cake}, \textit{That was a slice of cake}, \textit{That was easy}. We shall call these ``idiom strings'' ($I$), ``corrupted strings'' ($C$), and ``meaning strings'' ($M$), respectively. Next, we compute final-token cosine similarities between $M$ and all other strings (idiom and corrupted) $S$, for all intermediate layers $\ell$ of the network:
\begin{equation}
    \cos \theta_{M,S}^\ell = \frac{\vec x_M^\ell \cdot \vec x_S^\ell}{\|\vec x_M^\ell\| \|\vec x_S^\ell\|} \ ,
\end{equation}
where $\vec x_S^\ell$ is the final-token, layer $\ell$ residual stream from the forward pass of string $S$. We look at the final-token residual stream because only it has access to the full context of the string. For at least some $\ell$, $I$ should have a significantly higher similarity to $M$ compared to that between the $C$ and $M$, confirming that the model recognizes the idiomatic usage. 

\begin{figure}[ht]
\centering
\includegraphics[width=0.8\textwidth]{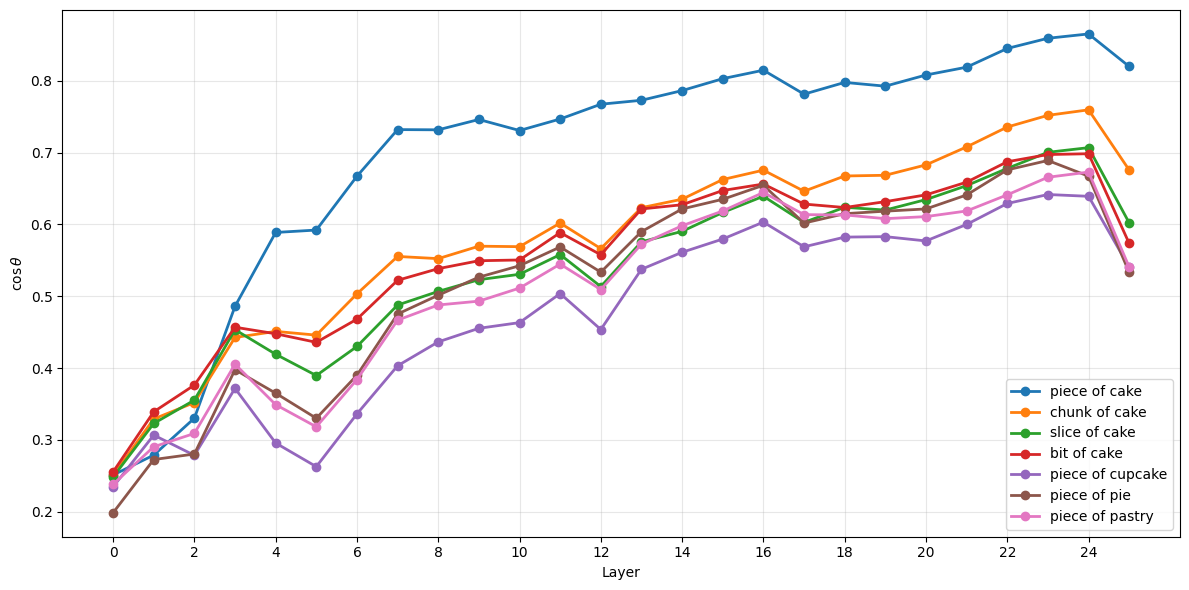}
\caption{Cosine similarity analysis for the idiom \textit{a piece of cake} and its corrupted variants, using the meaning string $M = \textit{That was easy}$. For each string $S = \textit{That was a ...}$, the cosine similarity between the final-token embedding vectors $\vec x_S^\ell$ and $\vec x_M^\ell$ is computed across all layers $\ell$. By layer 4, the idiom string is significantly more aligned with the meaning string than are the corrupted strings, demonstrating the effectiveness of the corruptions for isolating network components involved in its processing. This plot is representative of all idioms analyzed in this work (Table~\ref{tab:attention_effects}).}
\label{fig:cosine-sim}
\end{figure}

A representative example of this evaluation is shown in Figure~\ref{fig:cosine-sim}. Notice the similarity increase of \textit{a piece of cake} relative its corruptions in the early layers. By layer 4, the idiom shows a cosine similarity of $\sim 0.6$, while the corruptions remain below $\lesssim 0.45$.\footnote{Despite the high dimensionality (2304) of embedding space, the cosine similarities of even the corrupted strings are order one. This reflects the well-known fact that the internal representations of LLMs lie inside narrow cones (e.g., \citet{ethayarajh-2019-contextual}).} In later layers, increases in idiom similarity are accompanied by increases in corruption similarities, and represent only the changing geometry of embedding space. The interpretation is that by layer 4 the model has successfully integrated the idiomatic meaning into its internal representation. Note that we do not expect a similarity of $1$, given that, for example, in some contexts \textit{a piece of cake} might refer to an actual piece of cake. We shall denote by $L$ the layer after which the idiom's cosine similarity does not substantially increase relative to those of the corruptions (in this case $L=4$).

\subsection{Step 2: Single-corruption Circuit Discovery via Path Patching}

Once a list of corrupted strings has been generated, we treat them one at a time. Path patching \citep{wang2022interpretability, goldowsky2023localizing} can then be used to identify model components important for the processing of the idiom. The basic idea is to replace as many internal model activations in the forward pass of the idiom string as possible with those of the corrupted string, while still preserving the model's ability to represent the figurative meaning. The components that cannot be replaced without significant degradation of the intermediate layer representations are included in the discovered circuit. Our approach builds upon the Automated Circuit Discovery (ACDC) framework \cite{conmy2023acdc}, which we review below before commenting on several key methodological innovations that we introduced to provide further insights into transformer attention mechanisms.

\subsubsection{Review of the ACDC Algorithm}

The original ACDC algorithm \citep{conmy2023acdc} (also see \citet{hsu2024efficient} for recent work on circuit-building) operates through the following systematic process. One starts by representing the transformer, at some desired level of granularity, as a computational graph $G$. A coarser representation might contain only attention heads, while a finer one might include separate query, key, and value connections, and distinguish between tokens. The goal is to produce a minimal subgraph $H \subset G$ by initially setting $H = G$ and iteratively pruning unnecessary edges based on their causal importance.

First, order nodes from output to input (reverse topological order) to ensure that when evaluating a component's importance, all of its downstream effects have already been considered. For each node $v \in H$:
\begin{enumerate}
\item For each parent node $w$ that connects to $v$, temporarily remove the edge $w \to v$ from the current subgraph $H$. In practice, this involves replacing the activations from $w$ with those from the corrupted input during the forward pass.
\item Evaluate the performance drop. In the ACDC paper this is the change in the final-layer, final-logit Kullback-Leibler (KL) divergence: $d = D_{KL}(G || H_{new}) - D_{KL}(G || H)$.
\item If $d$ is less than a threshold $\tau$, permanently remove the edge from $H$. Otherwise, retain the edge as important for the circuit.
\end{enumerate}

The final subgraph $H$ represents the discovered circuit—a minimal set of components and connections sufficient to preserve the target behavior. The threshold $\tau$ controls the sparsity-performance tradeoff, with larger values producing sparser circuits.

\subsubsection{Custom Implementation for Idiom Processing}

\begin{figure}[ht]
\centering
\includegraphics[width=0.8\textwidth]{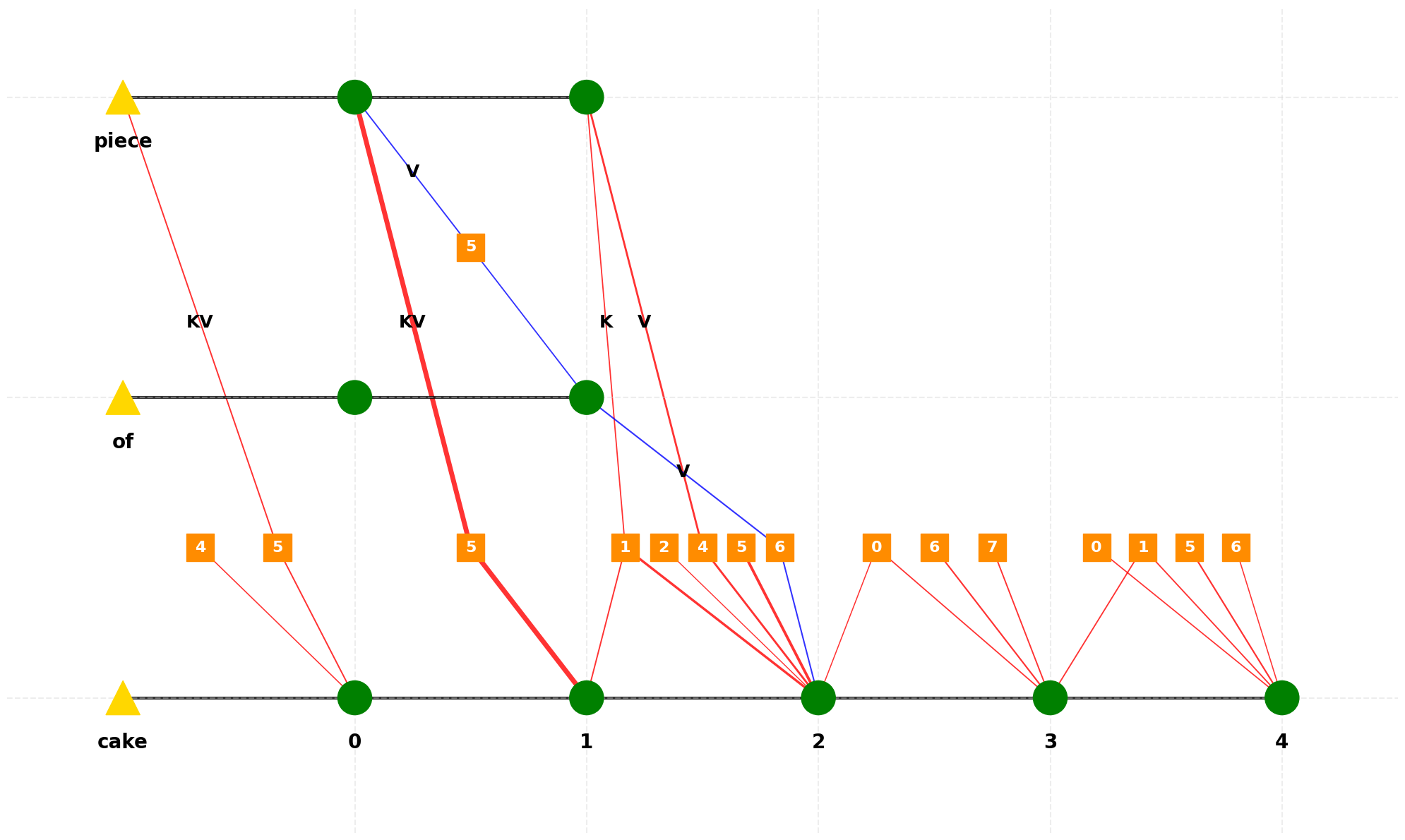}
\caption{A single-corruption ($\textit{piece}\to\textit{chunk}$) circuit discovered for the idiom \textit{a piece of cake} using $I = \textit{That was a piece of cake}$, $M = \textit{That was easy}$, $L = 4$, and $\tau = 0.005$. The yellow triangles represent embedding vectors, the green circles are the post-MLP residual stream, and the orange squares are attention heads (numbered 0--7 in Gemma 2--2B). Edges are colored red for performance drops and blue for gains (antagonistic components), with thickness proportional to the magnitude of the effect. Residual-to-residual edges are shown for completeness but are never patched. Cross-token edges are labeled K for Key, V for Value, and KV for both. The unambiguous Query edges are not labeled.}
\label{fig:single-circuit}
\end{figure}

Due to the unique nature of the problem at hand, we emphasize here several choices made in our implementation of the circuit discovery algorithm.

\textbf{Granularity of Computational Graph:} The nodes of the graph $G$ consist of all attention heads and post-MLP residual streams, for each token. This level of granularity is needed in order to track how specific tokens of the input attend to one another at each layer of the network. In addition to attention-head-to-residual edges within the same layer, there are also residual-to-attention-head edges of types Query (Q), Key (K), and Value (V) extending between adjacent layers (``incoming edges'').\footnote{The ability to independently patch Q/K/V connections between individual tokens makes this path patching, as distinguished from activation patching.} This allows us to identify the critical inputs to attention head behavior. Note that when Query edges are patched, we modify Q only for its attention to previous tokens, but not for its attention to itself (diagonal attention). Similarly, we do not include Key and Value edges between nodes of the same token. This is because, as mentioned at the beginning of this section, we are interested in the interactions between tokens, and not in the processing of information on individual tokens.

\textbf{Intermediate Layer Evaluation:} A fundamental difference in our approach is the choice of evaluation target. The original ACDC measures component importance by examining changes in the final output logits, the metric being the KL divergence between the full model ($G$) probability distribution and the pruned circuit ($H$) distribution. In contrast, our metric is the cosine similarity change in an intermediate layer $L$:
\begin{equation}\label{metric}
    \cos \theta_{M,H} = \frac{\vec x_M^L \cdot \vec x_{H}^L}{\|\vec x_M^L\| \|\vec x_{H}^L\|} \ ,
\end{equation}
where by $\vec x_{H}^L$ we mean the final-token, layer $L$ residual stream from the forward pass of $I$ but with the components $G \setminus H$ patched with the forward pass of $C$. In addition to reducing circuit discovery time by dropping layers beyond $L$, it is necessary to intercept the semantic meaning in middle of the network, rather than waiting for the final output where it may not be reflected in the next-token probability distribution. The performance drop is
\begin{equation}
    d = \cos \theta_{M,H} - \cos \theta_{M,H_{new}} \ .
\end{equation}

One could consider metrics other than \eqref{metric}. As just one example, one could train a sparse autoencoder (SAE) on the residual stream and compare instead the encoded representations of $\vec x_M^L$ and $\vec x_{H}^L$ (cosine similarity or otherwise). However, as shown in Figure~\ref{fig:cosine-sim}, the cosine similarity is sufficient to capture idioms' semantic processing.

\textbf{Inclusion of Antagonistic Components:} A final difference is that in contrast to ACDC, we retain circuit components whose patching leads to performance gains (in addition to performance drops) above threshold: $|d| > \tau$. This allows us to identify components that act to suppress the figurative interpretation of an idiom, though in practice such components are less common and without great effect.

A graphical representation of such a circuit is shown in Figure~\ref{fig:single-circuit}. Edges are weighted by their effect on circuit performance and colored: red for a performance drop and blue for a gain (antagonistic component). Note that while residual-to-residual edges are shown in the graph, these connections are never patched.

\subsection{Step 3: Threshold Determination}
\label{sec:threshold}

\begin{figure}[ht]
\centering
\includegraphics[width=0.8\textwidth]{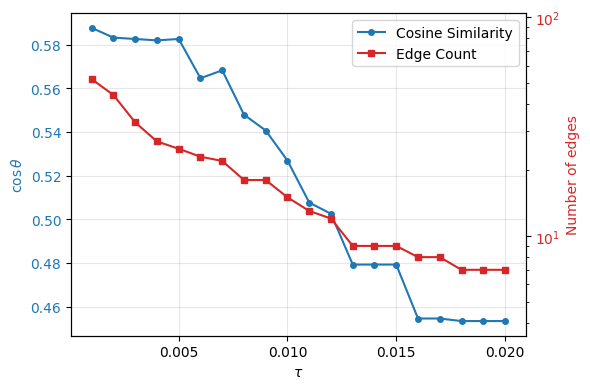}
\caption{Threshold sweep for the $\textit{cake}\to\textit{cupcake}$ corruption of \textit{a piece of cake} using $I = \textit{That was a piece of cake}$, $C = \textit{That was a piece of cupcake}$, $M = \textit{That was easy}$, and $L = 4$. The final-token cosine similarity between $C$ and $M$ (blue) and the log--number of edges (red) in the discovered circuit $H_C^\tau$ are plotted against the threshold $\tau$. In this case we would choose $\tau_* = 0.007$.}
\label{fig:sweep}
\end{figure}

Having constructed the circuit discovery algorithm, it is necessary to determine the optimal threshold $\tau_*$. A too-large value would miss critical elements of the circuit, while a too-small value would add components that do not contribute in general to idiom processing, but whose patching happens to be above threshold for the particular strings used. Therefore, we perform threshold sweeps - computing the number of edges and final cosine similarity score of $H^\tau$, over a range of values for $\tau$. The optimal threshold $\tau_*$ for each corrupted sentence should be greater than the range where the number of edges takes its final (low $\tau$) exponential behavior. On the other hand, $\tau_*$ should be below any significant jumps in the number of edges, indicating a large change in graph topology, likely due to the inclusion/exclusion of an important component. When in doubt, large jumps in the cosine similarity can be used to inform the decision. In practice, optimal thresholds generally fall in the range 0.004--0.008. An example sweep is plotted in Figure~\ref{fig:sweep}, for which we take $\tau_* = 0.007$.

\subsection{Step 4: Circuit Merging}

The final step is to take a number of circuits $H_C$ generated from single-token corruptions $C$, and merge them into one circuit $H_I$ representing the complete computational graph for the processing of an idiom. To do this we simply take the union of the nodes and edges from each circuit (each edge weight is computed as the maximum over the $H_C$). In addition to corrupting different tokens of the idiom, it is also important to corrupt each token in different ways. This is necessary because a given replacement token may not differ from the original token enough in some directions of embedding space to discover some circuit components. In this way, different corruptions are complementary. However, taken to an extreme this would lead to the inclusion of far too many components into the circuit, as the patching of any edge will likely exceed threshold for a different enough corruption (e.g., an unrelated non-English word). This is why we restrict to grammatically and (literally) semantically similar corruptions.

\begin{figure}[ht]
\centering
\includegraphics[width=0.8\textwidth]{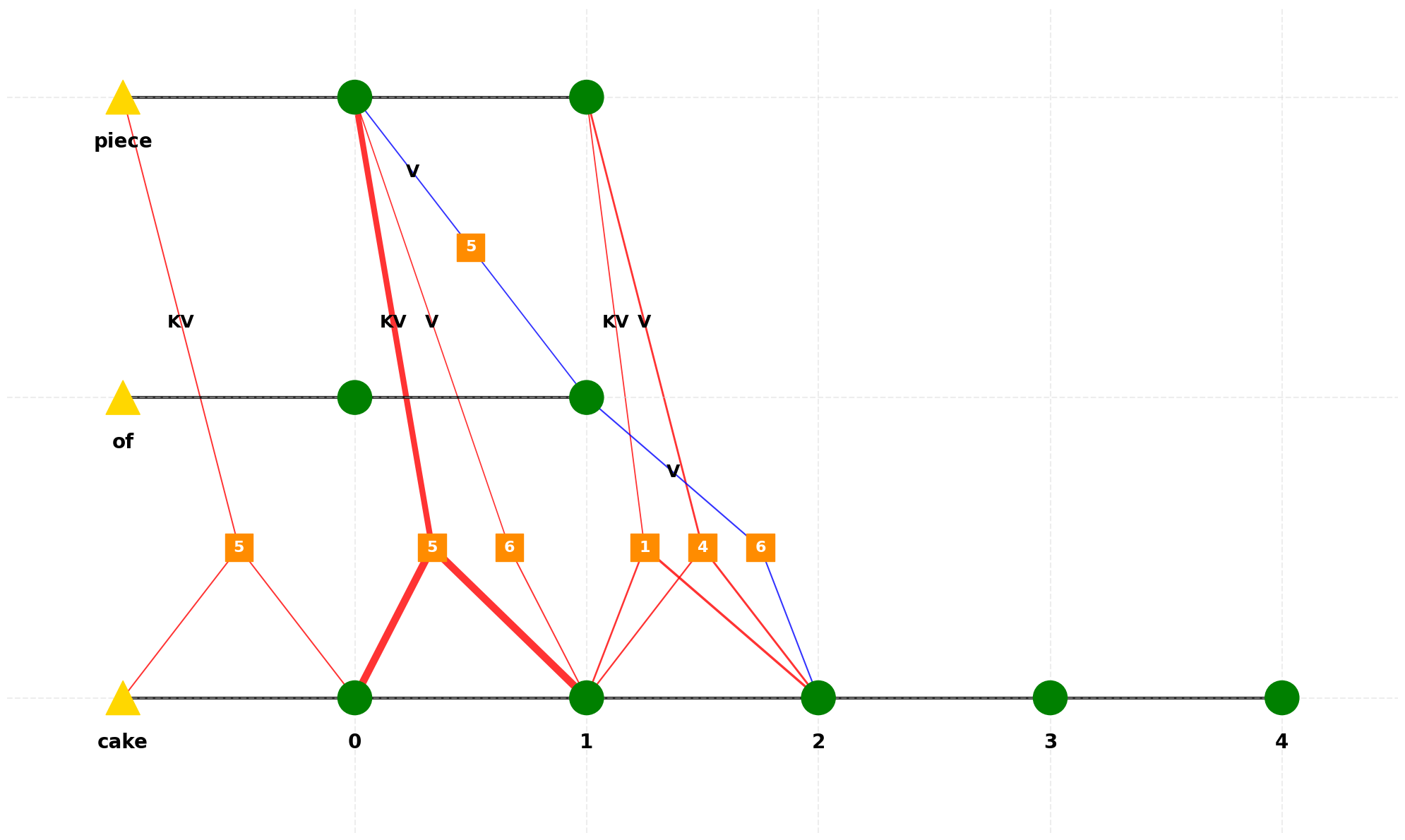}
\caption{The pruned circuit $\tilde H_I$ of the idiom \textit{a piece of cake} using $I = \textit{That was a piece of cake}$, $M = \textit{That was easy}$, and $L = 4$. The corruptions (thresholds) used are $\textit{piece}\to\textit{chunk/slice}$ (both $0.005$) and $\textit{cake}\to\textit{cupcake/pie}$ (both $0.007$). Notice how cross-token processing takes place in layers 0--2, before the idiom's figurative meaning is resolved in layers 3--4 (Figure~\ref{fig:cosine-sim}).}
\label{fig:merged}
\end{figure}

For the purpose of interpretability, once $H_I$ has been produced with the above algorithm, we make a final pruning of all attention heads that have either no incoming edges or only an incoming Query edge. Let us call the result $\tilde H_I$. The former case indicates that the head is acting like an MLP (single-token processing), while the latter signifies that the head is attending to a pre-idiom token in the string, which we are also not interested in. Note that in contrast to $H_I$, $\tilde H_I$, while more interpretable, will in general not faithfully reproduce $G$. An example is shown in Figure~\ref{fig:merged}.

\section{Results}
\label{sec:results}

Having established a circuit discovery methodology, we now present the results of its application to idiom processing in Gemma 2--2B. Our analysis reveals four key findings: (1) idiom processing circuits exhibit consistent structural properties across different expressions, with most computation occurring in early layers and final tokens; (2) specific attention heads consistently activate across multiple idioms, suggesting specialized ``Idiom Heads''; (3) the model employs distinct Query-Key space representations for each idiom rather than a universal idiom direction; and (4) early idiom processing creates ``augmented reception'' patterns that enhance later attention to idiom-specific tokens.

\subsection{General Circuit Properties}

Examination of the merged circuit presented in Figure~\ref{fig:merged}, obtained using corruptions of the tokens \textit{piece} and \textit{cake}, reveals several important features common to idiom processing circuits. First, all cross-token processing occurs in the earlier layers (0--2). Interestingly, the cosine similarity analysis of Figure~\ref{fig:cosine-sim} shows that the idiom string acquires no significant non-compositional meaning relative to the corruptions until layer 3. This can be understood using the idea of ``residual stream bandwidth'' \citep{elhage2021mathematical}: while in Gemma 2--2B the residual stream has dimension 2304, attention heads can only read from and write to 256-dimensional subspaces. Information written to the residual stream in early layers can therefore persist until it is ready to be processed. This is particularly important for idiom processing, where the model must maintain the possibility of both literal and figurative interpretations until sufficient context is available to disambiguate them.

In Table~\ref{tab:attention_effects} we present the attention heads that play a role in the processing of 8 common English idioms. While cross-token attention does play a role after layer 2 in some cases, the performance drops when patching these heads are generally insignificant. For simplicity only the two ``critical'' tokens of each idiom have been corrupted in the making of this table, leaving out function words like \textit{the} and \textit{of}. These tokens appear to play a less important role in idiom processing. As an example, attention to function tokens seems to be relatively unspecific, indicating a sharing of computational resources with circuits like ``\text{the} NOUN'' detection (see Appendix~\ref{app:function-words}). However, function tokens do play a role in ``augmented reception,'' discussed in Section~\ref{sec:augmented}.

Another feature of the discovered circuits is that most, if not all, of the idiomatic processing occurs on the final token. This can be seen in Figure~\ref{fig:merged}, in which only a single attention node on the token \textit{of} has been included in the circuit. In fact, of the 8 idioms analyzed in Table~\ref{tab:attention_effects}, this is the only example of such ``intermediate'' token processing. This suggests that the model defers computation until the very end of the idiom, when all necessary context is available - it makes no sense to waste resources performing anticipatory computations on \textit{piece of} when the next token may not be \textit{cake}.

There are also a few heads in Table~\ref{tab:attention_effects} that show antagonistic effects (negative performance drops). However, unlike the ``Negative Name Mover Heads'' of \citet{wang2022interpretability}, these heads are not consistently antagonistic and we therefore do not give them an analogous name. Furthermore, examination of the table shows that their effects are relatively small.

\begin{table}[!htb]
\centering
\small
\begin{tabular}{@{} l *{14}{c} @{}}
\toprule
& \multicolumn{2}{c}{Layer 0} & \multicolumn{4}{c}{Layer 1} & \multicolumn{3}{c}{Layer 2} & \multicolumn{4}{c}{Layer 3} & \multicolumn{1}{c}{Layer 4} \\
\cmidrule(lr){2-3} \cmidrule(lr){4-7} \cmidrule(lr){8-10} \cmidrule(lr){11-14} \cmidrule(lr){15-15}
Idiom            & H4& H5  & H1& H2& H3& H5  & H0& H1& H4  & H0& H3& H5& H6  & H2 \\
\midrule
\textit{\textbf{kicked} the \textbf{bucket}} & 7* & 9   & 5* & --& 5 & 14  & 20*& --& --  & --& --& --& --  & --\\
\textit{a \textbf{piece} of \textbf{cake}}  & --& --  & --& --& --& 15  & --& 2 & 2   & --& --& --& --  & --\\
\textit{\textbf{hit} the \textbf{sack}}     & --& -1*  & --& --& 2* & --  & 15& -1& -1  & --& --& --& --  & --\\
\textit{\textbf{rocket} \textbf{science}}   & --& --  & --& 3 & --& 2   & --& --& --  & --& 1 & --& 1   & --\\
\textit{\textbf{missed} the \textbf{boat}}  & 1 & --  & --& 3 & --& --  & 9 & --& 1   & --& --& --& 2   & 3 \\
\textit{\textbf{pulling} your \textbf{leg}} & --& --  & --& 3 & 3 & --  & 11& -2& --  & --& --& --& -1  & 2 \\
\textit{\textbf{under} the \textbf{weather}}& 1 & --  & --& 16& --& --  & 14& --& 5   & --& --& 1 & --  & --\\
\textit{\textbf{faced} the \textbf{music}}   & --& --  & --& --& 9 & --  & 9 & --& --  & 1 & --& --& --  & 3 \\

\bottomrule
\end{tabular}
\caption{Attention head performance drops $d$ in units of $10^{-2}$ for the circuits of 8 common idioms, corrupting only the two bolded tokens of each. Negative values indicate antagonistic effects that suppress idiomatic interpretation. Only drops with $|d|$ > 0.01 are shown. Asterisks indicate heads without incoming Query edges for the corruptions tested. All parameters used in the circuit discovery can be found in Appendix~\ref{app:params}.}
\label{tab:attention_effects}
\end{table}

\subsection{Identification of Idiom Heads}

One of the most striking features of Table~\ref{tab:attention_effects} is the recurrence of certain attention heads exhibiting large performance drops across multiple idioms, which we dub ``Idiom Heads.'' A clear example is head 0 of layer 2, which we write as $(2,0)$. The heads $(1,2)$, $(1,3)$, and $(1,5)$ are also strong candidates. These heads nearly always have incoming Query edges, in addition to Key and/or Value edges, indicating that they are idiom-specific. The few exceptions (marked with asterisks in the table) indicate that, for example in the case of \textit{kicked the bucket}, only the less specific ``bucket-ness'' of \textit{buckets} and \textit{pail} is needed to trigger head $(2,0)$'s attention to \textit{kicked}. We suppose that, due to the multiple heads involved in the processing of each idiom, such non-specificity is acceptable.

In fact, the presence of multiple heads in each row of Table~\ref{tab:attention_effects} exemplifies a common feature of transformer circuits—redundancy—which arises from the tension between computational efficiency and polysemanticity. Indeed, with only 8 attention heads per layer, each must take on many roles. While having just one head for idiom processing would be more efficient, it would also make the model less robust to different contexts in which, for example, that head was needed for another task. On the other hand, given that each head only writes to a small subspace of the residual stream, if we presume that the idiomatic information stored in preparation for single-token processing also resides in a small subspace, it makes sense that only a few heads should play a role.

\subsection{Exploring QK Space}

The existence of Idiom Heads implies the sharing of computational resources across different idioms. Where does this sharing occur? One possibility is on the subspaces on the residual stream written to by these heads, as discussed above. Another possibility is in the heads' Query-Key (QK) space. That is, is there a universal idiom direction in QK space along which idiom-participating tokens have a sizable component? To test this, we collect the Query and Key vectors of head $(2,0)$ from the critical (original and corrupted) tokens of the 4 syntactically identical idioms \textit{kicked the bucket}, \textit{hit the sack}, \textit{missed the boat}, and \textit{faced the music}. Their dot products are shown in Table~\ref{tab:qk-dot}.

In most cases, the non-diagonal entries are less than even the corrupted dot products in parentheses (see table caption for how to read these entries). In the few cases where this is not true, the non-diagonal products are still significantly smaller than the diagonal (idiomatic) ones. Given the exponential sensitivity (due to softmax) of attention weights to QK dot products, we can conclude that there is no universal idiom direction in QK space. Instead, each idiom has its own unique direction, shared between its critical tokens.

\begin{table}[!htb]
\centering
\small
\begin{tabular}{@{} l c c c c @{}}
\toprule
 & \textit{kicked} & \textit{hit} & \textit{missed} & \textit{faced} \\
\midrule
\textit{bucket} & 72 (66, 28) & 50 & 10 & 22 \\
\textit{sack}   & 14 & 81 (50, 34) & 25 & 44 \\
\textit{boat}  & 18 & 6 & 70 (30, 29) & 8 \\
\textit{music} & 8 & 40 & 0 & 81 (-3, 27) \\
\bottomrule
\end{tabular}
\caption{Dot products between Query (rows) and Key (columns) vectors of head $(2,0)$ for the critical tokens of the idioms \textit{kicked the bucket}, \textit{hit the sack}, \textit{missed the boat}, and \textit{faced the music}. For a fair comparison we use the template $S = \textit{He \underline{\hspace{2em}} the \underline{\hspace{2em}}}$. In parentheses are, respectively, the averaged dot product between corrupted Queries and the original Key (compare to same column) and the averaged dot product between the original Query and corrupted Keys (compare to same row).}
\label{tab:qk-dot}
\end{table}

\subsection{Augmented Reception}
\label{sec:augmented}

When the attention node of a non-final-token, single-corruption circuit has an incoming Query edge, we call this ``augmented reception.'' An example can be seen in Figure~\ref{fig:single-circuit}, where the attention node $(2,1)$ on the final token \textit{cake} has both incoming Query and Key edges. This demonstrates that idiomatic processing in earlier layers informs the attention patterns of later tokens and can be understood as a mechanism by which the model more efficiently allocates computational resources: by attending less to \textit{slice} in \textit{a slice of cake}, attention can instead be given to other possibly more important tokens in the context.

\begin{figure}[ht]
\centering
\includegraphics[width=0.49\textwidth]{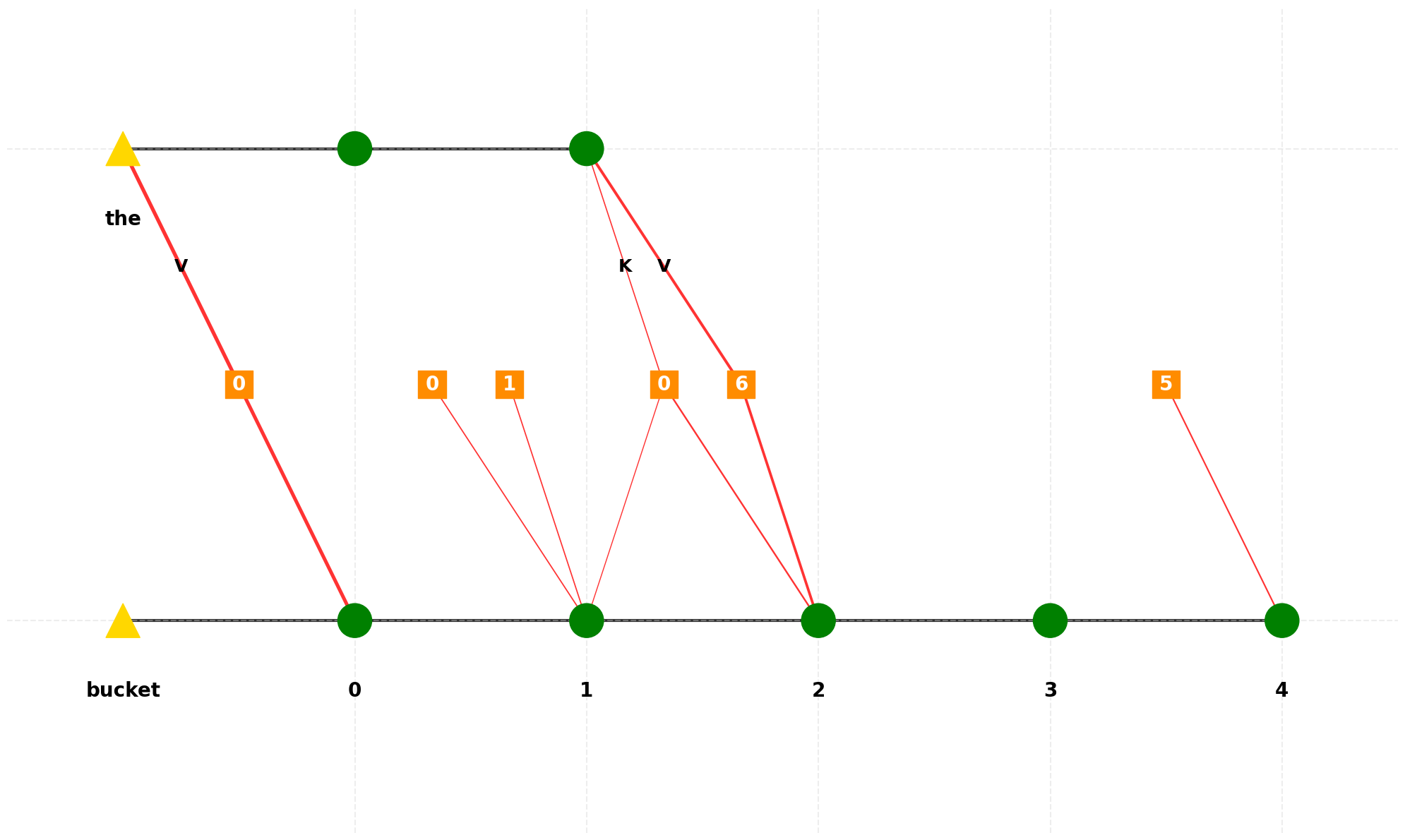}
\includegraphics[width=0.49\textwidth]{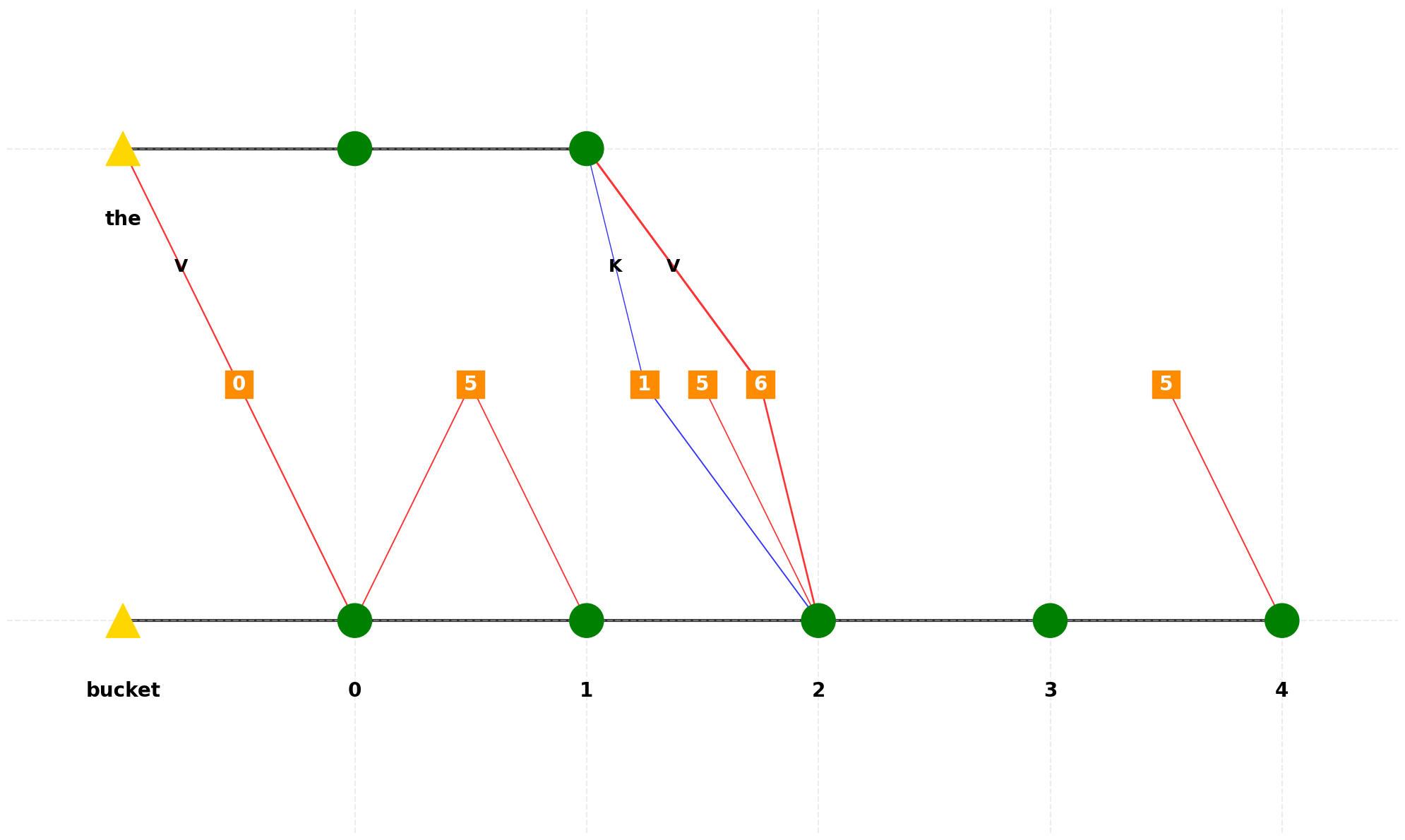}
\caption{Circuits for the idiom \textit{kicked the bucket} using $I = \textit{He kicked the bucket}$, $M = \textit{He died}$, and $L = 4$. The corruptions (thresholds) used are $\textit{the}\to\textit{a}$ ($0.010$) and $\textit{the}\to\textit{this}$ ($0.009$). Notice how attention heads $(2,0)$ and $(1,5)$, respectively, on the final token \textit{bucket} have incoming Query edges, demonstrating augmented reception.}
\label{fig:augmented}
\end{figure}

Another role played by augmented reception is in the interaction of non-final tokens. Due to the lack of attention nodes on non-final tokens (recall that the single such node in Figure~\ref{fig:merged} is an exception), one might presume that these interactions only occur in single-token processing (MLP layers or diagonal/pre-idiom attention). However, a non-final token can also influence an attention head's receptivity to others. As an example, see the two circuits in Figure~\ref{fig:augmented}, for the corruption of $\textit{the}\to\textit{a/this}$ in \textit{kicked the bucket}. This phenomenon can be observed in nodes $(2,0)$ and $(1,5)$, respectively, both of which attend strongly to \textit{kicked} (Table~\ref{tab:attention_effects}).

\section{Discussion}
\label{sec:discussion}

Our findings demonstrate that transformers like Gemma 2--2B process the non-compositional meaning of idioms in their early layers (0--5). The processing over the course of multiple layers reflects the flexibility required due to the ambiguous nature of language, maintaining both literal and figurative interpretations until sufficient context is available, and aligning well with psycholinguistic theories of idiom comprehension \citep{cacciari1983idioms, titone2008idiom}. Furthermore, the sequential nature of idiom processing—with cross-token computation in earlier layers (0--2) and semantic integration later (layers 3--5)—aligns with the idea of ``residual stream bandwidth'' \citep{elhage2021mathematical} and enhances robustness to different contexts. Early layers establish potential interpretative pathways that persist until sufficient context enables disambiguation. Moreover, the deferral of idiomatic processing until the final token represents an efficient use of computational resources.

The identification of ``Idiom Heads'' provides further evidence for functional specialization within the transformer attention mechanism. These heads exhibit frequent activation across diverse idioms, paralleling previous discoveries of Induction Heads \citep{olsson2022context} and Duplicate Token Heads \citep{wang2022interpretability}. However, the idiom-specific nature of Query-Key space representations entails an important caveat: Idiom Heads encode idiom-specific directions rather than a universal ``idiom direction.'' The redundancy across multiple heads reflects the tension between efficiency and robustness, providing protection against changing contexts while maintaining specialized functionality. Perhaps our most novel finding, augmented reception demonstrates how early idiom processing modifies the receptivity patterns of later attention heads. That is, the model actively modulates the allocation of attention based on expectations from earlier processing. This mechanism may help to explain how transformers are able to synthesize information over long context lengths.

A limitation of our analysis is that it focuses on English idioms in a single model architecture. While none of the circuit phenomena we discovered appear to depend upon these specifics, broader study of more idioms in multiple languages and transformer models is needed to confirm their generality. Furthermore, threshold determination introduces subjective elements. A more systematic approach might try to quantify the size of ``non-idiomatic noise'' in patching. An interesting direction for future work would be to investigate whether similar patterns occur for other non-compositional phenomena, like grammatical constructions. Additionally, further examples of and investigation into augmented reception should lead to a deeper understanding of transformer information processing.

\section*{Acknowledgments}

The author thanks Boris Barron and Vassilis Papadopoulos for many fruitful discussions and for their feedback on the manuscript.

\bibliographystyle{plainnat}
\bibliography{references}

\newpage

\appendix

\section{Code Use and Availability}
\label{sec:appendix-code}

This work uses TransformerLens \citep{nanda2022transformerlens}. Our implementation is available at:

\noindent \url{https://github.com/agomes42/LLMConstructions}

\section{Candidate Corruptions by Cosine Similarity}
\label{app:corruptions}

\begin{table}[H]
\centering
\small
\begin{tabular}{@{} r r c r @{\hspace{5em}} r r c r @{}}
\toprule
 & Token ID & String & $\cos \theta$ &  & Token ID & String & $\cos \theta$ \\
\midrule
1 & 6431 & \_piece & 1.0000 & 21 & 98113 & \_pezzi & 0.3718 \\
2 & 17143 & piece & 0.7959 & 22 & 52352 & piec & 0.3710 \\
3 & 9308 & \_pieces & 0.7638 & 23 & 104739 & \_peça & 0.3543 \\
4 & 38669 & \_Piece & 0.7489 & 24 & 176208 & \textjapanese{ピース} & 0.3487 \\
5 & 36992 & Piece & 0.7231 & 25 & 31951 & \_pièces & 0.3428 \\
6 & 163978 & \_PIECE & 0.7213 & 26 & 141263 & \_Stücke & 0.3320 \\
7 & 40192 & pieces & 0.6204 & 27 & 124603 & \_morceaux & 0.3214 \\
8 & 66337 & \_Pieces & 0.5879 & 28 & 120984 & stücke & 0.3191 \\
9 & 94628 & Pieces & 0.5575 & \textbf{29} & \textbf{34657} & \textbf{\_chunk} & \textbf{0.3183} \\
10 & 62800 & \_pieza & 0.5344 & 30 & 127827 & \_parça & 0.3137 \\
11 & 206148 & PIECE & 0.5160 & 31 & 43428 & \_peças & 0.3104 \\
12 & 150685 & \_pezzo & 0.4538 & 32 & 160626 & ECES & 0.2916 \\
13 & 33382 & \_piezas & 0.4366 & 33 & 7124 & \_peace & 0.2907 \\
14 & 46756 & \_pièce & 0.4249 & 34 & 140282 & \_stück & 0.2900 \\
15 & 150985 & \_morceau & 0.4171 & 35 & 146364 & \_peda & 0.2849 \\
16 & 109633 & \_piec & 0.4051 & 36 & 19763 & \_fragment & 0.2847 \\
17 & 54344 & \_Stück & 0.4046 & 37 & 17447 & \_bits & 0.2790 \\
18 & 72334 & \_stuk & 0.3940 & 38 & 127947 & stuk & 0.2783 \\
19 & 47338 & stück & 0.3829 & \textbf{39} & \textbf{27125} & \textbf{\_slice} & \textbf{0.2752} \\
20 & 181216 & {\selectlanguage{russian} \_кусо} & 0.3766 & 40 & 18233 & iece & 0.2685 \\
\bottomrule
\end{tabular}
\caption{The 40 most similar tokens (Gemma 2 tokenizer) to \textit{piece} by embedding space cosine similarity, showing potential corruptions for circuit discovery. An underscore indicates a space character in the token. The most suitable corruptions in the idiom \textit{a piece of cake} are \textit{chunk} and \textit{slice}, shown in bold.}
\label{tab:corruptions}
\end{table}

\section{Function Words and Idiom Processing}
\label{app:function-words}

\begin{figure}[H]
\centering
\includegraphics[width=0.8\textwidth]{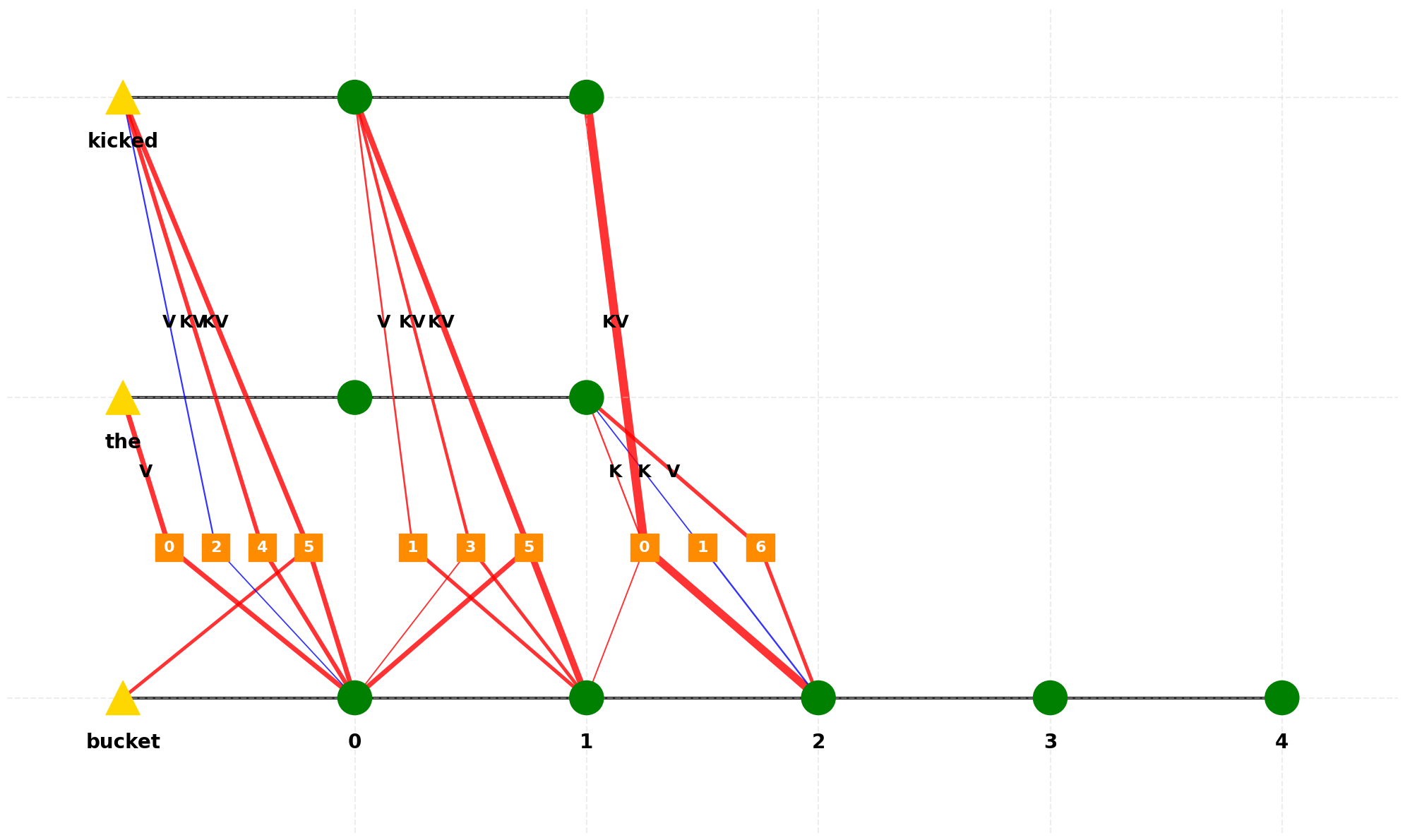}
\caption{A circuit for the idiom \textit{kicked the bucket} using $I = \textit{He kicked the bucket}$, $M = \textit{He died}$, and $L = 4$. The corruptions (thresholds) used are $\textit{kicked}\to\textit{booted/punched}$ ($0.008$/$0.010$), $\textit{the}\to\textit{a/this}$ ($0.010$/$0.009$), and $\textit{bucket}\to\textit{buckets/pail}$ ($0.006$/$0.009$). Notice how attention heads $(0,0)$ and $(2,6)$ have no incoming Query edges, indicating that rather than being idiom-specific, they are performing ``\textit{the} NOUN'' detection.}
\label{fig:function-words}
\end{figure}

\section{Circuit Discovery Parameters}
\label{app:params}

\begin{table}[H]
\centering
\small
\begin{tabular}{@{} l *{4}{l} @{}}
\toprule
Idiom string           & Corruption 1& Corruption 2& Corruption 3& Corruption 4 \\
\midrule
\textit{He \textbf{kicked} the \textbf{bucket}} & \textit{booted} (0.008) & \textit{punched} (0.010) & \textit{buckets} (0.006) & \textit{pail} (0.009) \\
\textit{That was a \textbf{piece} of \textbf{cake}}  & \textit{chunk} (0.005) & \textit{slice} (0.005) & \textit{cupcake} (0.007) & \textit{pie} (0.007) \\
\textit{He \textbf{hit} the \textbf{sack}}     & \textit{struck} (0.005) & \textit{shot} (0.009) & \textit{sacks} (0.006) & \textit{bag} (0.004) \\
\textit{It's not \textbf{rocket} \textbf{science}}  & \textit{missile} (0.005) & \textit{spaceship} (0.008) & \textit{scientist} (0.005) & \textit{physics} (0.007) \\
\textit{You \textbf{missed} the \textbf{boat}}  & \textit{skipped} (0.003) & \textit{lost} (0.005) & \textit{boats} (0.004) & \textit{sailboat} (0.007) \\
\textit{I'm \textbf{pulling} your \textbf{leg}} & \textit{pushing} (0.003) & \textit{grabbing} (0.004) & \textit{legs} (0.004) & \textit{arm} (0.003) \\
\textit{She is \textbf{under} the \textbf{weather}}& \textit{in} (0.004) & \textit{over} (0.004) & \textit{climate} (0.005) & \textit{rain} (0.007) \\
\textit{He \textbf{faced} the \textbf{music}}   & \textit{confronted} (0.009) & \textit{encountered} (0.007) & \textit{musical} (0.006) & \textit{songs} (0.008) \\

\bottomrule
\end{tabular}
\caption{The corruptions (thresholds) used to generate Table~\ref{tab:attention_effects}. Corruptions 1 and 2 replace the first bolded token, while Corruptions 3 and 4 replace the second bolded token. The meaning strings $M$ (final layer $L$) used are, in order: \textit{He died} (4), \textit{That was easy} (4), \textit{He went to sleep} (5), \textit{It's not difficult} (4), \textit{You missed the opportunity} (4), \textit{I'm kidding you} (4), \textit{She is feeling ill} (5), and \textit{He faced the consequences} (5).}
\label{tab:circuit_params}
\end{table}

\end{document}